\documentclass[runningheads]{llncs}
\usepackage[T1]{fontenc}
\usepackage{amsmath}
\usepackage{amssymb}
\usepackage{comment}

\usepackage{subcaption}
\usepackage{placeins}

\usepackage{graphicx}
\usepackage{subcaption}
\begin{document}
%

\title{Does the Model Say What the Data Says? A Simple Heuristic for Model–Data Alignment}
\titlerunning{Does the Model Say What the Data Says?}
%
\author{Henry Salgado\inst{1}\orcidID{0009-0001-6506-5484} \and
Meagan R. Kendall\inst{2}\orcidID{0000-0002-9940-4405} \and
Martine Ceberio\inst{1}\orcidID{0000-0001-5680-1155}}
\authorrunning{H. Salgado et al.}
%
\institute{Department of Computer Science, The University of Texas at El Paso, El Paso, TX, USA \and
Department of Engineering Education and Leadership, The University of Texas at El Paso, El Paso, TX, USA}
\maketitle              
\begin{abstract}
 In this work, we propose a simple, computationally efficient framework to evaluate whether machine learning models align with the structure of the data they learn from, that is, whether \textit{the model says what the data says}. Unlike existing interpretability methods that focus exclusively on explaining model behavior, our approach establishes a baseline derived directly from the data itself. Drawing inspiration from Rubin's Potential Outcomes Framework, we measure how strongly each feature separates two outcome groups in a binary classification task, going beyond traditional descriptive statistics to quantify each feature's effect on the outcome. By comparing these data-derived feature rankings against model-based explanations, we provide practitioners with an interpretable method to assess model–data alignment.
 

\end{abstract}
%
%
\vspace*{-.3cm}
\section{Introduction}

As Deep Learning models grow in power and complexity, they have also become increasingly opaque. For example, modern architectures such as those used by Large Language Models (LLMs) often contain billions or even trillions of parameters, effectively turning them into ``black boxes'' whose internal mappings between weights and decisions are largely uninterpretable to humans~\cite{lee_deep_2024}. This lack of transparency poses serious challenges in high-stakes domains where understanding model reasoning is critical in establishing trust, accountability, and fairness~\cite{molnar_interpretable_2020}.

The consequences of this interpretability gap are already evident across multiple application areas. In healthcare, image-based diagnostic systems have been shown to rely on spurious correlations, such as associating the presence of medical equipment (e.g., chest drains or tubes) with disease severity rather than actual pathological features~\cite{zech_variable_2018}~\cite{rueckel_pneumothorax_2021}. In criminal justice, widely deployed risk-assessment tools such as COMPAS have exhibited biases that disproportionately disadvantage certain demographic groups, leading to inequitable sentencing and parole outcomes~\cite{christian_alignment_2020}.
Even in the domain of reasoning LLMs, models sometimes produce chain-of-thought outputs that appear logical but contain flawed reasoning steps~\cite{chen_reasoning_2025}.

These examples highlight the pressing need for research on interpretability methods, causal reasoning, and data quality, three interconnected areas that, together, could help practitioners determine how much to trust model outputs. In this work, we propose a simple, computationally efficient framework to evaluate whether trained models align with the structure of the data they learn from. Unlike existing interpretability methods that focus exclusively on explaining model behavior, our approach establishes a baseline derived directly from the data itself. We measure how strongly each feature separates two outcome groups in a binary classification task, going beyond traditional descriptive statistics (e.g., means, mode) to quantify each feature's effect size on the outcome. By comparing these data-derived feature rankings against model-based explanations, we provide practitioners with an interpretable approach to assess model–data alignment.


We demonstrate our approach on two widely available datasets, comparing our method against feature importance from a decision tree and SHapley Additive exPlanations (SHAP) values from a neural network. Preliminary experiments yield promising results, showing that all three methods consistently identify the same top features.

\vspace*{-.1cm}
\section{Related Work}

\vspace*{-.1cm}
\subsection{White-box Models and Post-hoc Techniques}

Researchers have generally taken two main approaches to improve interpretability. The first is to use simpler, more transparent models, such as linear regression, where the relationships between variables can be traced directly. The second approach employs post-hoc explainability methods, such as Local Interpretable Model-agnostic Explanations (LIME) or SHAP (SHapley Additive exPlanations)~\cite{ribeiro_why_2016}, to provide explanations for black-box model predictions.

While both strategies have produced valuable insights, they remain limited in important ways. White-box models depend on simplifying assumptions such as linearity and normality that often fail in high-dimensional, real-world data. Post-hoc methods, meanwhile, are computationally expensive and typically highlight associations rather than causal relationships, which can mislead users about the true drivers of model outputs~\cite{molnar_general_2022}.

\vspace*{-.1cm}
\subsection{Causal Methods}

A growing body of work has begun to integrate causal reasoning into machine learning to move from correlation to causation. In natural language processing, researchers have examined how specific linguistic properties causally influence outcomes such as sentiment or classification~\cite{pryzant_causal_2021}. In computer vision, others have explored whether learned features correspond to the same causal features used by experts, such as dermatologists identifying skin lesions~\cite{vedaldi_determining_2020}. There is also increasing interest in evaluating whether LLMs can answer causal questions and perform causal discovery on previously unseen data~\cite{zecevic_causal_2023}.

While these approaches have produced promising results, they also face several challenges. In many cases, the underlying causal graph is unknown, confounding is difficult to address, and the computational costs of these methods are high. These issues make it challenging to scale causal approaches to large datasets or real-world applications.

\vspace*{-.1cm}
\subsection{Data Explainability}

Most existing interpretability approaches, whether white-box, post-hoc, or causal, focus on explaining model behavior given a dataset. While this focus is important, it can overshadow another fundamental issue: the quality and explainability of the data itself. A model's reliability depends heavily on the data it learns from. Researchers have long emphasized the importance of using data from reliable sources, handling missingness properly, and checking for distributional imbalances before model training~\cite{mohammed_effects_2025}.

For example, in the case of missing data, traditional methods such as Rubin's Missing Completely at Random (MCAR) framework~\cite{rubin_multiple_2004} provide structured ways to handle incomplete information. Practitioners also rely on descriptive statistics and visualization techniques to understand data distributions and identify outliers~\cite{hatcher_advanced_2013}. In this vein, we also argue that careful attention to data understanding should be treated as a key step in improving transparency and trust in machine learning.

\vspace*{-.1cm}
\section{Methodology}

To advance this line of reasoning, we draw inspiration from Rubin's Potential Outcomes Framework~\cite{rubin_estimating_1974} and adapt it to evaluate model fidelity relative to the data it learns from. The goal is to determine whether \textit{the model says what the data says}. Using this framework, every observation has potential outcomes under certain conditions. For example, a passenger on the Titanic could either have survived or not survived depending on their conditions.

Causal effects are formally defined as the difference between these potential outcomes. However, this difference cannot be directly observed for any single individual, since only one outcome is realized. The strength of the causal framework lies in comparing randomized groups. For example, groups \(A\) and \(B\), where the difference in average outcomes can be interpreted as the Average Causal Effect (ACE).
\begin{equation}
    \text{ACE} = \mathbb{E}[Y(1)] - \mathbb{E}[Y(0)].
\end{equation}
Here, \(Y(1)\) represents the outcome that would be observed if an individual (or unit) were exposed to a condition or treatment (e.g., survival if rescued), while \(Y(0)\) represents the outcome that would occur in the absence of that condition (e.g., not surviving if not rescued). The causal effect for an individual is the difference \(Y(1) - Y(0)\), and the ACE represents this difference averaged across all individuals in the population.\\


Our proposed approach follows a similar logic. We consider a binary classification problem, one in which there are two possible outcomes. Each outcome is treated as a group. For each feature in each group, we calculate the mean and variance, then compute the difference between the two group means. These differences are standardized by the pooled standard deviation to account for differences in feature units. The result is a Standardized Mean Difference (SMD), which quantifies the magnitude of separation between the two groups for each feature. 

Formally, let the dataset be: $\mathcal{D} = \{(x_i, y_i)\}_{i=1}^{n}$,
where \(x_i = (x_{i1}, x_{i2}, \ldots, x_{ip}) \in \mathbb{R}^p\) and \(y_i \in \{0, 1\}\). 
We define: $\mathcal{D}_1 = \{x_i : y_i = 1\}$,  
    $\mathcal{D}_0 = \{x_i : y_i = 0\}$.
For each feature \(j\),
\begin{equation}
    \mu_{1j} = \frac{1}{n_1} \sum_{x_i \in \mathcal{D}_1} x_{ij}, \quad
    \mu_{0j} = \frac{1}{n_0} \sum_{x_i \in \mathcal{D}_0} x_{ij},
\end{equation}
\begin{equation}
    s_{p,j} = \sqrt{\frac{(n_1 - 1)s_{1j}^2 + (n_0 - 1)s_{0j}^2}{n_1 + n_0 - 2}},
    \quad
    \Delta_j = \frac{\mu_{1j} - \mu_{0j}}{s_{p,j}}.
\end{equation}
The absolute value \(|\Delta_j|\) represents the standardized effect size of feature \(j\). We rank features by descending \(|\Delta_j|\), which provides an ordering of features by their discriminative power.


This approach offers two advantages over the existing methods. First, by analyzing the data directly rather than model outputs, we establish an independent baseline that reveals what patterns are present before the model is trained. Comparing this baseline to post-hoc explainability method outputs, whether the model has learned to prioritize the same features that statistically distinguish classes in raw data. This comparison provides practitioners with another tool to gain confidence in their findings. Second, the method is computationally efficient, requiring only basic statistical calculations rather than permutations of features needed for most popular post-hoc explainability or causal methods.

\vspace*{-.1cm}

\section{Evaluation}
\subsection{Datasets}
We evaluated our approach using two established binary classification datasets to demonstrate generalizability across different data and feature types:

\subsubsection{Titanic Dataset}
The Titanic dataset contains information about 891 passengers with a binary outcome variable, \textit{Survived}, where \(y = 1\) indicates survival and \(y = 0\) indicates death. The dataset includes both continuous and categorical features:
\begin{itemize}
    \item \textbf{Pclass:} Passenger class (1st, 2nd, 3rd)
    \item \textbf{Sex:} Male or female 
    \item \textbf{Age:} Passenger age in years
    \item \textbf{SibSp:} Number of siblings or spouses aboard
    \item \textbf{Parch:} Number of parents or children aboard
    \item \textbf{Fare:} Ticket fare paid
    \item \textbf{Embarked:} Port of embarkation (C = Cherbourg, Q = Queenstown, S = Southampton)
\end{itemize}

\subsubsection{Pima Indians Diabetes Dataset}
The Pima Indians Diabetes dataset contains medical measurements for 768 female patients with a binary outcome variable, \textit{Outcome}, where \(y = 1\) indicates diabetes diagnosis and \(y = 0\) indicates no diabetes. All features are continuous:
\begin{itemize}
    \item \textbf{Pregnancies:} Number of pregnancies
    \item \textbf{Glucose:} Plasma glucose concentration
    \item \textbf{BloodPressure:} Diastolic blood pressure (mm Hg)
    \item \textbf{SkinThickness:} Triceps skin fold thickness (mm)
    \item \textbf{Insulin:} 2-hour serum insulin (mu U/ml)
    \item \textbf{BMI:} Body mass index (weight in kg/(height in m)$^2$)
    \item \textbf{DiabetesPedigreeFunction:} Diabetes pedigree function
    \item \textbf{Age:} Age in years
\end{itemize}

\subsection{Data Preprocessing and Experimental Setup}
For both datasets, we applied consistent preprocessing:
\begin{itemize}
    \item \textbf{Missing values:} Imputed instances using median (Diabetes: no missing values)
    \item \textbf{Numerical encoding:} For categorical features (Titanic only)
    \item \textbf{Feature scaling:} StandardScaler normalization for neural network models
    \item \textbf{Train-test split:} 80\%-20\% split with fixed random seed (random\_state=42)
\end{itemize}

\subsection{Model Configurations}
To evaluate model-data alignment, we trained two model types on each dataset using identical hyperparameters across datasets for consistency.

\subsubsection{Neural Network (Multi-Layer Perceptron)}
We implemented a feedforward neural network using scikit-learn's MLPClassifier with the following configuration:
\begin{itemize}
    \item \textbf{Architecture:} hidden\_layer\_sizes=(128, 64, 16)
    \item \textbf{Activation:} ReLU
    \item \textbf{Optimizer:} Adam with learning\_rate\_init=0.001
    \item \textbf{Training:} max\_iter=400, early\_stopping=True, validation\_fraction=0.1
    \item \textbf{Random seed:} random\_state=42
\end{itemize}
Features were standardized using StandardScaler before training. We computed SHAP values using KernelExplainer with 100 background samples to estimate feature contributions to predictions.

\subsubsection{Decision Tree}
We implemented a decision tree classifier using scikit-learn with the following configuration:
\begin{itemize}
    \item \textbf{Maximum depth:} max\_depth=5
    \item \textbf{Split criterion:} criterion='entropy'
    \item \textbf{Minimum samples to split:} min\_samples\_split=15
    \item \textbf{Minimum samples per leaf:} min\_samples\_leaf=10
    \item \textbf{Random seed:} random\_state=42
\end{itemize}
We extracted feature importances based on information gain, which measures each feature's contribution to reducing impurity across all splits.

\subsection{Alignment Metrics}
To quantitatively assess model-data alignment, we computed rank correlation metrics between data-level feature importance (standardized mean differences) and model-level feature importance (decision tree importance and SHAP values):

\begin{itemize}
    \item \textbf{Spearman's rank correlation ($\rho$):} Measures the relationship between rankings, values range from -1 to 1, where values $>0.7$ indicate strong agreement.
\end{itemize}

Additionally, we visualized alignment through scatter plots comparing feature rankings across methods, with the diagonal representing perfect agreement.

\vspace*{-.1cm}

\section{Results and Discussion}
\subsection{Model Performance}
Both models achieved reasonable classification performance on both datasets. Table~\ref{tab:model_performance} shows training and testing accuracies, indicating that the models successfully learned predictive patterns without severe overfitting.

\begin{table}[!htbp]
\centering
\caption{Model Performance on Test Sets}
\label{tab:model_performance}
\begin{tabular}{lcccc}
\hline
\textbf{Model} & \multicolumn{2}{c}{\textbf{Titanic}} & \multicolumn{2}{c}{\textbf{Diabetes}} \\
 & Train & Test & Train & Test \\
\hline
Neural Network & 0.813 & 0.783 & 0.789 & 0.727  \\
Decision Tree  & 0.841 & 0.776 & 0.822 & 0.779  \\
\hline
\end{tabular}
\end{table}

\FloatBarrier

\subsection{Quantitative Model–Data Alignment}

To quantitatively assess whether models learn patterns aligned with the data structure, we computed rank correlations between data-level feature importance (SMD) and model-level importance measures (Decision Tree importances and SHAP values). Table~\ref{tab:alignment_metrics} presents Spearman’s rank correlation coefficients for both datasets.

\begin{table}[h!]
\centering
\caption{Model–Data Alignment: Spearman’s Rank Correlation ($\rho$). Higher values indicate stronger alignment between feature importance rankings.}
\label{tab:alignment_metrics}
\vspace{0.2cm}
\begin{tabular}{lcc}
\hline
\textbf{Method Comparison} & \textbf{Titanic} & \textbf{Diabetes} \\
\hline
SMD vs. Decision Tree & 0.642 & 0.761\\
SMD vs. SHAP & 0.607 & 0.952 \\
\hline
\end{tabular}
\end{table}

\textbf{Titanic Dataset:}
The rank correlation between SMD–SHAP revealed moderate alignment ($\rho = 0.607$), as did SMD–Decision Tree ($\rho = 0.642$). This weaker alignment may reflect that the model is capturing non-linear interactions (e.g., between age, sex, and passenger class) that are not evident in univariate SMD analysis.

\textbf{Diabetes Dataset:}
Both methods showed strong agreement, with high correlation between SMD and Decision Tree importances ($\rho = 0.761$) and especially between SMD and SHAP values ($\rho = 0.952$).

\subsection{Detailed Rank Comparison via Scatter Plots}

\begin{figure}[h!]
\centering
\captionsetup[subfigure]{font=scriptsize,labelfont=bf}

\textbf{(A) Titanic Dataset}\\
\begin{subfigure}{0.45\textwidth}
    \includegraphics[width=\textwidth]{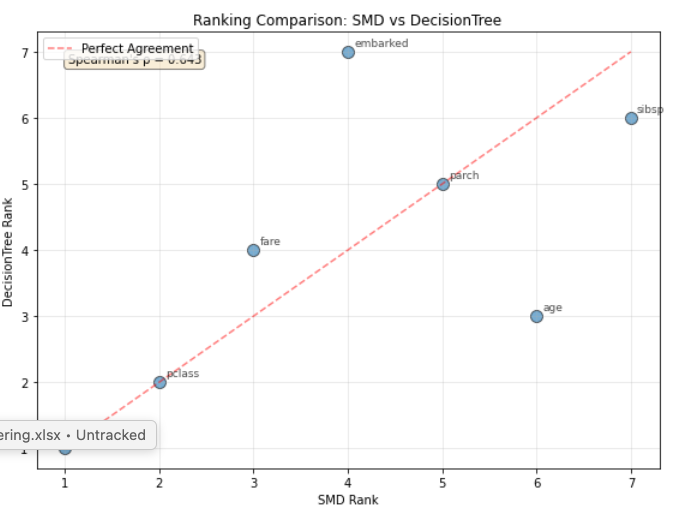}
    \caption{SMD vs. Decision Tree ($\rho=0.642$)}
\end{subfigure}
\hfill
\begin{subfigure}{0.45\textwidth}
    \includegraphics[width=\textwidth]{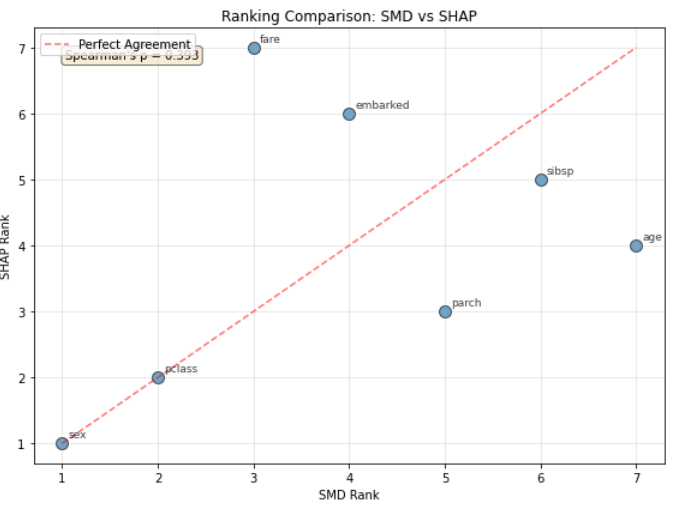}
    \caption{SMD vs. SHAP ($\rho=0.607$)}
\end{subfigure}

\caption{Rank comparison scatter plots for the Titanic dataset. Points close to the diagonal indicate agreement between the feature ranking methods.}
\label{fig:scatter_titanic}
\end{figure}

\begin{figure}[h!]
\centering
\captionsetup[subfigure]{font=scriptsize,labelfont=bf}

\textbf{(B) Diabetes Dataset}\\
\begin{subfigure}{0.45\textwidth}
    \includegraphics[width=\textwidth]{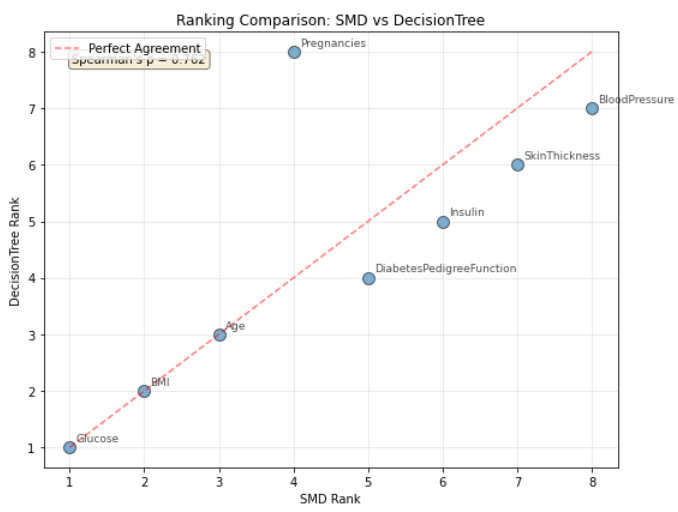}
    \caption{SMD vs. Decision Tree ($\rho=0.761$)}
\end{subfigure}
\hfill
\begin{subfigure}{0.45\textwidth}
    \includegraphics[width=\textwidth]{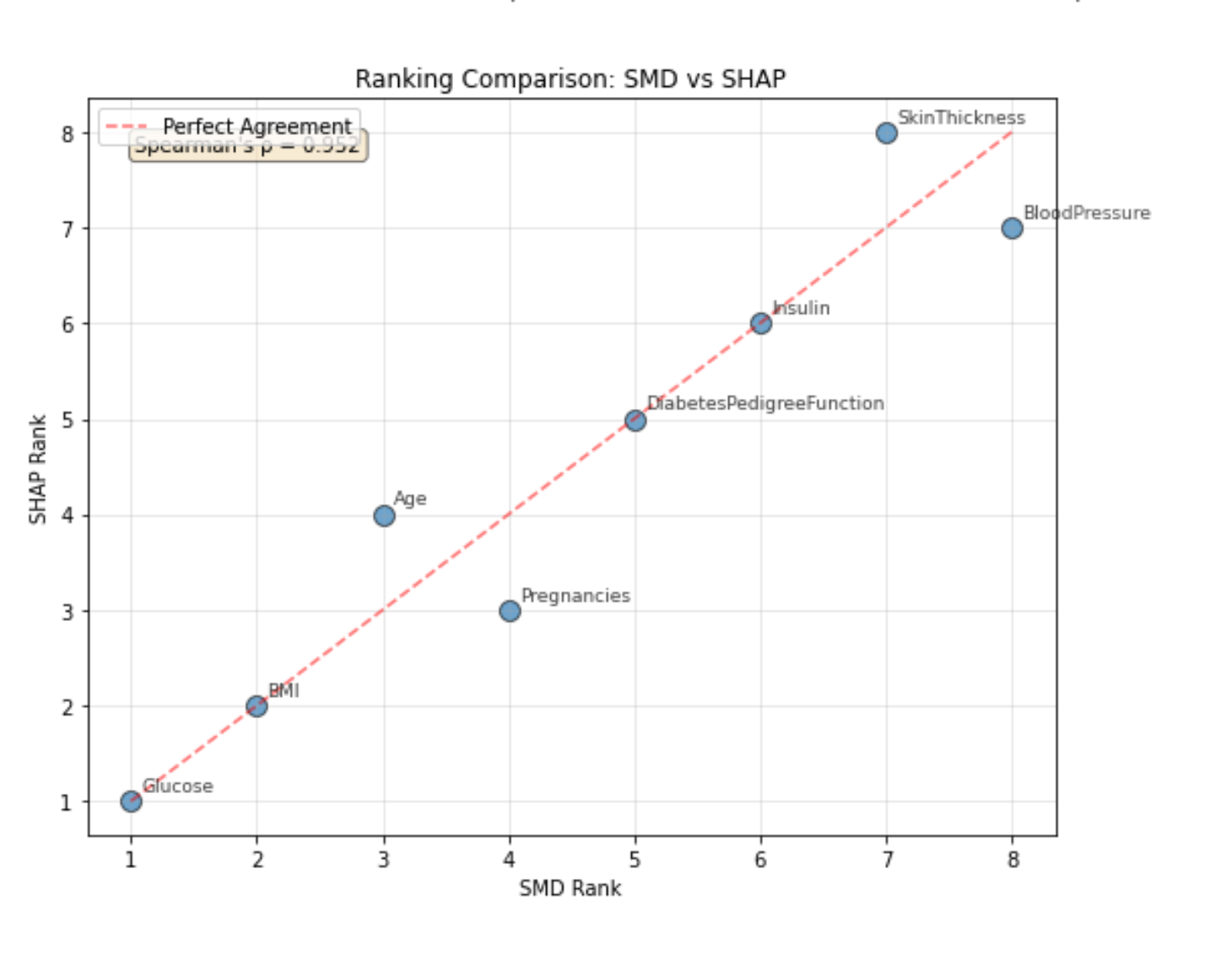}
    \caption{SMD vs. SHAP ($\rho=0.952$)}
\end{subfigure}

\caption{Rank comparison scatter plots for the Diabetes dataset. Both comparisons show strong diagonal clustering, indicating high alignment between SMD-derived feature rankings and model-based importance measures.}
\label{fig:scatter_diabetes}
\end{figure}

To visualize the agreement in feature rankings, Figures~\ref{fig:scatter_titanic} and \ref{fig:scatter_diabetes} present scatter plots comparing SMD-derived ranks to those from Decision Tree importances and SHAP values.

\FloatBarrier

\FloatBarrier

\textbf{Titanic Dataset:}
The SMD–SHAP and SMD–Decision Tree scatter plots both show moderate alignment, with most points lying near the diagonal and a handful of features deviating more noticeably. This pattern is consistent with their similar rank correlations (SMD–SHAP, $\rho = 0.607$; SMD–Decision Tree, $\rho = 0.642$), indicating that all three methods broadly agree on which features matter most. Minor differences in the ordering of lower-ranked features likely reflect how each model class (tree-based vs.\ neural network) captures non-linearities and interactions that are not visible in univariate SMD. Key predictors such as \textit{Sex} and \textit{Class} remain highly ranked across methods.

\textbf{Diabetes Dataset:}
Both scatter plots exhibit strong diagonal alignment, consistent with high correlations between SMD and Decision Tree importances ($\rho = 0.761$) and especially between SMD and SHAP values ($\rho = 0.952$). This consistent clustering indicates robust agreement across methods and confirms that dominant predictors (e.g., \textit{Glucose}, \textit{BMI}) are identified similarly by both statistical and model-based approaches.
\vspace*{-.2cm}

\section{Conclusion, Limitations, and Future Work}

In this work, we introduced a simple and computationally efficient data–model alignment heuristic for binary classification tasks. We do note that rather than serving as a causal inference method, this approach provides a quantitative means of assessing whether model-derived feature importance aligns with data-driven statistical patterns. By comparing feature rankings obtained directly from data (e.g., standardized mean differences) with those derived from model explanations (e.g., decision tree importances, SHAP values), the method complements existing interpretability techniques by focusing on the consistency between model learning and data structure. While the current framework is limited to binary classification and does not yet extend to multiclass or regression problems, it offers a scalable diagnostic for model–data coherence. Notably, the heuristic may flag features as important even when their effects are mediated or confounded by other variables, underscoring its descriptive rather than causal intent. Future work should expand this framework to handle multi-class and continuous outcomes, integrate more robust statistical tests for alignment, and explore approaches to distinguish direct from mediated feature effects.

\bibliographystyle{unsrt}  
\bibliography{references}  

\end{document}